\documentclass{article}
\usepackage{arxiv}
\usepackage[utf8]{inputenc} 
\usepackage[T1]{fontenc}    
\usepackage{hyperref}       
\usepackage{url}      

\usepackage{breakurl}
\usepackage{booktabs}       
\usepackage{amsfonts}       
\usepackage{nicefrac}       
\usepackage{microtype}      
\usepackage{lipsum}
\usepackage{authblk}
\usepackage{comment}
\usepackage{multicol}
\usepackage{csquotes}
\usepackage{graphicx}
\usepackage{float}
\usepackage{caption}
\usepackage{subcaption}
\newcommand*{\email}[1]{%
    \normalsize\href{mailto:#1}{#1}\par
    }

\title{(Un)Masked COVID-19 Trends from Social Media}

\author[1]{Asmit Kumar Singh\textsuperscript{*}\thanks{These authors contributed equally to this paper}}

\author[1]{Paras Mehan\textsuperscript{*}}
\author[2]{Divyanshu Sharma\textsuperscript{*}}
\author[3]{Rohan Pandey\textsuperscript{*}}
\author[1]{Tavpritesh Sethi\thanks{Correspondence at: \email{tavpriteshsethi@iiitd.ac.in}}}
\author[1]{Ponnurangam Kumaraguru}
\affil[1]{Indraprastha Institute of Information Technology, Delhi}
\affil[2]{Netaji Subhas University of Technology, Delhi}
\affil[3]{Shiv Nadar University, Noida}

\date{}

\begin{document}
\maketitle

\begin{abstract}
The adoption of non-pharmaceutical interventions and their surveillance is critical for detecting and stopping possible transmission routes of COVID-19. A study of the effects of these interventions in terms of adoption can help shape public health decisions. Also, the efficacy of Non-Pharmaceutical Interventions can be affected by public behaviours in events such as election rallies, festivals and protest events, as captured from social media. Social media analytics can offer crucial public health insights Here we examined mask use and mask fit in the United States, especially  during the first large scale public gathering post pandemic, the Black Lives Matter (BLM). This study aimed to analyze the utilization and fit of face masks and social distancing in the USA from social media and events of large physical gatherings through publicly available social media images from six cities and the BLM protests. 2.04 million publicly available social media images were collected and analyzed from the six cities between February 1, 2020, and May 31, 2020. We used correlation tests to examine the relationships between the online mask usage trends and the COVID-19 cases. We looked for significant changes in mask-wearing patterns and group posting before and after important policy decisions. 
For BLM protests, we analyze 195,452 posts from New York and Minneapolis from May 25, 2020, to July 15, 2020. We looked at differences in adopting the preventive measures in the BLM protests through the mask-fit score. The average percentage of group pictures dropped from 8.05\% to 4.65\% post the lockdown week. New York City, Dallas, Seattle, New Orleans, Boston and Minneapolis observed an increase of 5\%, 7.4\%, 7.4\%, 6.5\%, 5.6\% and 7.1\% in mask wearing online, respectively, between February 2020 and May 2020. Boston and Minneapolis observed a significant increase of 3\% and 7.4\% mask-wearing after the mask mandates. A difference of 6.2\% and 8.3\% were found in the group pictures between BLM posts and Non-BLM posts for New York City and Minneapolis. In contrast, the difference between BLM and NON-BLM posts in the percentage of masked faces in group pictures was 29\% and 20.1\% for New York City and Minneapolis, respectively. Of the masked faces in protests, 35\% wore the mask with a fit score greater than 80\%. The study finds a significant drop in the group posting when the stay-at-home laws were applied and a significant increase in mask wearing for two of the three cities when the mask mandates were applied. Although a general positive trend towards mask-wearing and social distancing is observed, a high percentage of posts did not adhere to the guidelines. BLM related posts were found to capture the lack of seriousness to safety measures through a high percentage of group pictures and low mask fit scores. Thus, the methodology used provides a directional indication of how government policies can be indirectly monitored through social media. 

\end{abstract}

\section{Introduction}
The outbreak of COVID-19 has the world in its grips. The World Health Organisation (WHO) declared it as a global pandemic on March 11, 2020 \cite{worldhealthorganization_2020_who}, and with exponentially rising cases, there are currently more than 32 million cases and 500,000 deaths(April 2021) \cite{a2020_johns}. Following WHO health advisories, most countries have declared national emergencies, closed borders, and restricted public movement \cite{nossiter_2020_restrictions,petersen_2020_covid19}. Masks have been found to reduce potential exposure risk from an infected person, proving to be a successful measure to suppress transmission and save lives \cite{chu_2020_physical,howard_2021_an,rader_2021_maskwearing,clapham_2021_face}. Many studies have recognized the importance of community-wide wearing masks for controlling the pandemic \cite{cheng_2020_the,worby_2020_face}. Social distancing measures were also applied to prevent sick individuals from coming in contact with healthy individuals. These social distancing measures and mask-wearing have proven successful in many countries like China\cite{ainslie_2020_evidence,du_2020_effects}. Governments worldwide have used social distancing and mask-wearing as a primary non-pharmaceutical measure against the virus.

With over 32 million cases and 500,000 deaths(as of April, 2021), the USA  is one of the largest hit countries by the virus. In the USA, many state governments had applied several stay-at-home and mask-wearing measures as early non-pharmaceutical interventions. In the lead-up to widespread vaccine deployment, the adoption of non-pharmaceutical interventions and their surveillance is critical for detecting and stopping possible transmission routes. Quantifying the effectiveness of such measures is a challenging task, which currently relies on on-ground surveys \cite{haischer_2020_who} or self-reported numbers\cite{rader_2021_maskwearing}. However, these methods are cumbersome, thus leading to lags in data and day-to-day evolution of a fast moving pandemic. 

The pervasive nature of social media provides a unique opportunity to create agile frameworks for assessing public health measures such as mask use. With its ease of access and global outreach, social media has a disproportionate influence on disseminating information during a pandemic \cite{depoux_2020_the}. In recent times of the pandemic, social media has become a popular platform for people to express their thoughts, opinions, and broadcast activities. The general public and authorities have been using hashtags like \#CoronaOutbreak, \#COVID19, and \#mask to disseminate important information and health advisories, which provides us with an opportunity to analyze behaviours and the impact of such advisories worldwide \cite{a2021_covid19}. Indeed, social media has been extensively explored and analyzed for patterns that have emerged during the COVID-19 pandemic \cite{ahmad_2020_the} \cite{cinelli_2020_the,huang_2020_mining,islam_2020_covid19related}. Signorini et al. \cite{signorini_2011_the} examined Twitter based-information to track the swiftly-evolving public sentiment regarding Swine Flu in 2011 and correlate the H1N1 virus subtype-related activity to track reported disease levels in the US accurately. 

During the pandemic, the country also observed the Black Lives Matter protests. The killing of George Floyd on 25th May 2020 sparked a series of protests\cite{taylor_2020_george} and agitation across the country. Protests are designed to stimulate public action for social justice. Such protests involve the physical gathering of people, making it difficult to adhere to social distancing and to wear masks. These protests provide an opportunity to observe how people react to public health-related preventive measures during such gatherings, but collecting the necessary data from the ground is a really hard task. Such protests have also gained high popularity and attraction through online social media \cite{mundt_2018_scaling,cox_2017_the}

Realizing the potential of social media in understanding such events, in this study, we utilize the social media images from Instagram, a popular image-sharing social media platform. Computer vision-based classifiers are necessary to check if a person is wearing a mask from its image. There exist two datasets previously published for mask classification tasks, MAFA(MAsked FAces)\cite{ge_2017_detecting} and RMFD(Real world Masked Face Dataset)\cite{jiang_2020_retinamask}. MAFA dataset contains 35,805 masked images. Since the MAFA Dataset was curated and released in 2017, it couldn’t capture different varieties and types of masks that have been in use during the pandemic period. The MAFA dataset is biased towards one kind of mask; it majorly consists of medical staff wearing disposable medical-grade masks. RMFD contains 7,959 masked images containing a variety of masks used during the COVID-19 period. However, manual qualitative evaluation of the images revealed that the images were not suitable for analyzing high-quality social media images since the majority of the images were of less than 50x50 resolution after cropping their face region. In addition to mask detection, analyzing the fit of the mask is a highly useful application. There is no previous work that is trying to analyze mask fit using semantic segmentation and a corresponding annotated dataset for the same to the best of our knowledge.   

Therefore, this study fills the gap with a pipeline designed to estimate the extent of mask behaviors by assessing mask use and mask fit from 2.04 million social media images from six US cities (Table 1). Along with the geographical diversity among the cities, the six cities also have high population numbers. These cities were also found to have a high number of location-tagged posts on Instagram and hence were chosen as the locations of interest. We demonstrate the correlation of mask use and mask fit behaviors with COVID-19 burden, policy directives and large scale events such as the nationwide BLM protest in these six cities.  

\section{Methods}
\subsection{Datasets}
This article uses three different dataset collections for different purposes. 
\begin{enumerate}
    \item Mask-Unmask Classifier Dataset : This dataset is used for training a model that classifies whether the person in the image is wearing a mask or not. So we needed images in which people were wearing masks (masked images) and images in which people were not wearing masks (unmasked images) so that our model can learn to distinguish between the two categories. We collected around 30,000 images of people wearing masks from Google Search Images using tags "people wearing masks", "children wearing masks". Images from Instagram were also collected with the Tag explore feature, using these three tags: "mask", "masked", "covidmask". Although the tagging algorithms used by Google is expected to capture most of the images, some images may have been potentially missed. There is further scope for expanding our set of tags chosen to capture the entire population of images in which people are wearing masks, and this is a limitation of our current approach. However, this will need more research as capturing other scenarios may also lead to noisier sets. Post data collection, images with a width and height of at least 50 pixels were kept to ensure decent image quality. Then Face Detector was used on these images to extract faces. The images of extracted faces were distributed between 5 annotators, and annotators were asked to classify the face as either "masked" or "unmasked". After the annotations, images of 9,055 masked faces were obtained. This dataset of 9,055 masked images (VAriety MAsks - Classification (VAMA-C)) is publicly available for use in further research\cite{precog}. For un-masked face images, we made a random sample(without replacement) of size 9,055 from the VGGFace2 Dataset \cite{cao_2018_vggface2}, which is a large-scale face recognition dataset. Sample from VGGFace2, along with VAMA-C (2.1), was used to train the Unmask-Mask Classifier, whose details are given in the Proposed Frameworks section. 
    \item Fit Score Dataset: Out of 9,055 masked faces that we obtained from the previous dataset, we selected 504 images with different poses and a wide variety of mask designs. Then we annotated these images using Label Studio \cite{label} for getting pixel-level annotations of mask region on the face. This dataset was then used to train a semantic-segmentation model, whose details are in Proposed Frameworks section. This dataset (VAriety MAsks - Segmentation (VAMA-S)) is publicly available for use in further research\cite{precog}.
    \item USA-Cities Instagram dataset: For the analysis phase, we collected the location-tagged public posts from Instagram between February 1, 2020, and May 31, 2020, for six cities New York City, Seattle, Dallas, New Orleans, Minneapolis, and Boston. The first covid case was reported in January 2020 \cite{cnn_us}, and till July 2020 US is still in the first wave of the coronavirus pandemic \cite{mukaka_2012_statistics}. Hence, the chosen time frame captures the beginning and growth of the US's COVID-19 pandemic. This collection was done for six major US cities. These six cities were selected to represent different geographical sections of the country. We collected a total of 2.04 million public posts from these six cities. These posts were collected via Instagram's explore location feature. Instagram's GraphQL API was employed for the data collection. The tools used have been published as a python PyPI package\cite{insta_scraper}. The states and the summary statistics can be found in the (Table 1). We also collected data for New York City and Minneapolis from May 25, 2020, to July 15, 2020. These two cities observed major protests\cite{a2020_some}. We curate a list of trending tags and keywords  (‘blm’,  ‘blacklivesmatter’, ‘ georgefloyd’, ‘justiceforgeorgefloyd’, ‘policebrutality’, and ‘protest’ ) during this period. We refer to the posts whose captions included these tags as BLM posts and the rest as Non-BLM posts. 
\end{enumerate}

\subsection{Proposed Framework}
This article proposes a Mask-Unmask classification framework (for classifying masked and unmasked images) and a fit score analysis framework (for evaluating whether the masks are being worn effectively or not from the given image). The Mask-Unmask classification model is used to analyze the "USA-Cities Instagram dataset" and The Fit-score analysis framework is just used for BLM posts, in order to capture the mask-wearing patterns during a huge social gathering.

Images obtained from sources mentioned in the previous section consisted of various individuals. Thereby, to detect face masks in these images, both the framework's first task was face detection. This was done using the pre-trained model Retinaface\cite{deng_2019_retinaface}, which is one of the top-performing models on Face Detection on the WIDER Face(Hard) dataset\cite{yang_2016_wider}. Next, facial landmarks were obtained using the Dlib's implementation\cite{king_2009_dlibml} proposed by Kazemi et al.\cite{kazemi_2014_one}. The landmark nos. 5-13, 31-36, 49-68 (Figure \ref{fig:fig5}) were used to filter the face's jaw region. This jaw region obtained was then used as input to the classification model for the mask-unmask classification framework. The landmark nos. 32-36, 49-68 (Figure \ref{fig:fig5}) was used to filter the nose-mouth region from the face. This nose-mouth region obtained was then used for calculating the fit-score. 

\begin{figure}[ht!]
	\centering
	\includegraphics[width=0.95\linewidth]{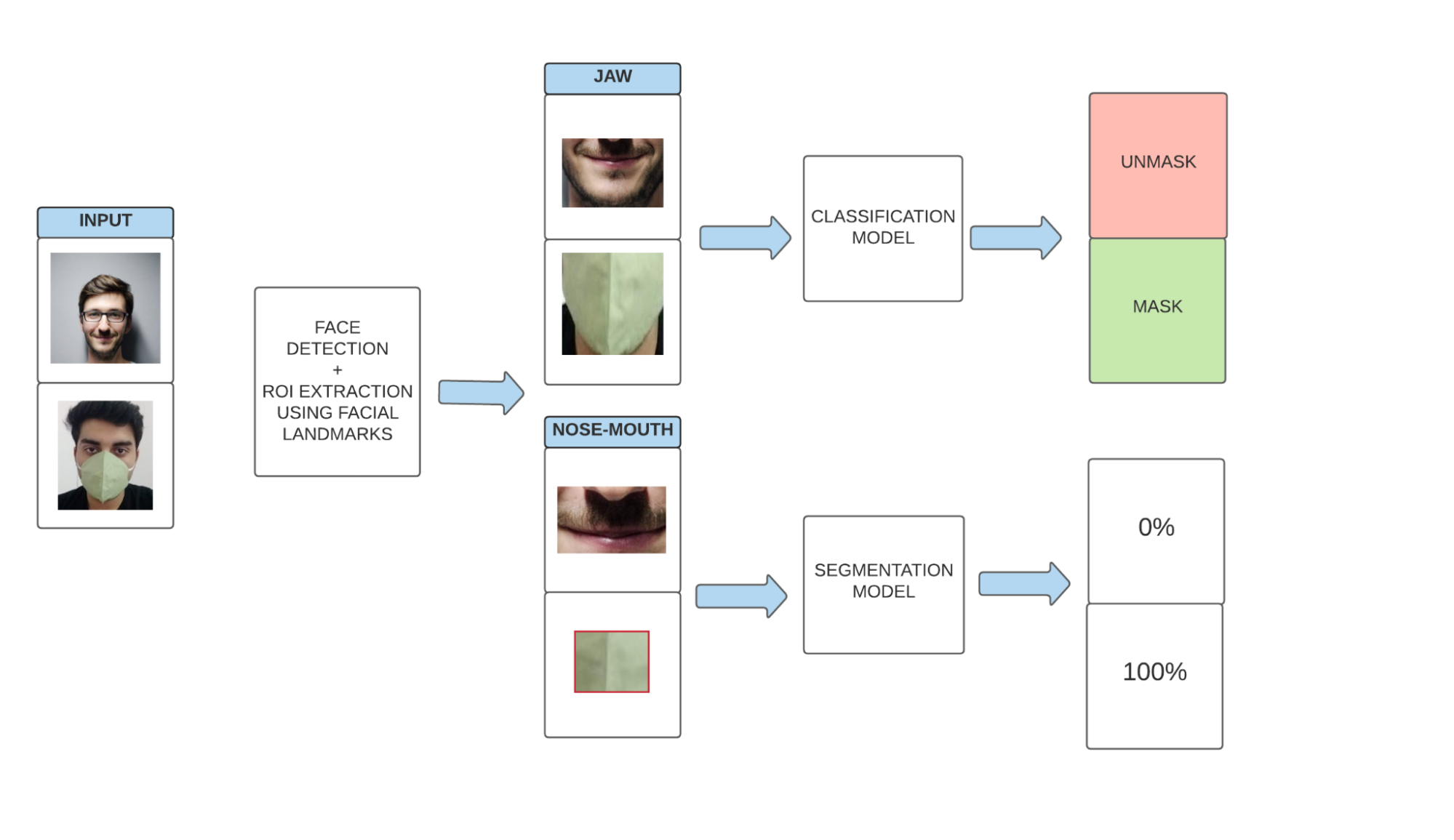}
	\caption{Face mask detection, and mask fit calculation framework . The input image is passed through a face detection model, from which the Region of Interests (ROIs) are extracted using facial landmarks. The extracted jaws using facial landmarks, are then passed to the trained mask-unmask classification model. While the extracted nose-mouth region, is passed to the segmentation model to predict the masked region, and calculate the fit score.}
	
	\label{fig:fig1}
\end{figure}

After obtaining the image of the jaw, the images were classified depending on whether the jaw region contained a mask over it or not (Figure \ref{fig:fig1}). The following architectures were experimented with while training a mask-unmask classification model - MobileNet V2, Nas Net, EffecientNet B0, EffecientNet B1, EffecientNet B2, DenseNet121. These models were selected since they have significantly fewer parameters than most other architectures (Table 12). 
The input image size for all the models was 224 x 224. Adam optimizer with an initial learning rate of 1e-4 was used, and each model was trained for 30 epochs with a batch size of 64, with binary cross-entropy as the loss function. For training the classifier for detecting mask and unmask from the input jaw regions, the total data from Section 2.1 consisted of 9,055 masked and 9,055 unmasked samples, which were split into 80:20 ratio of train and validation set. 5-Fold Cross-validation was used to evaluate the trained models' performance, and the results are shown in (Table 12). 

The evaluation indicates that EfficientNet B0 emerges as the best performing model with an overall accuracy of 0.98 ± 0.01. EfficientNet \cite{tan_2019_efficientnet} is a convolutional neural network architecture and scaling method that uniformly scales all depth/width/resolution dimensions. Efficient B0 has 5.3 Million parameters with 18 layers. Its architecture consists of an initial 3x3 convolution layer, followed by a series of MBconv layers with different kernel sizes and no. of channels. The series of MBconv layers are followed by a convolution layer, a pooling layer, and a fully connected layer. It uses linear activation in the last layer in each block to prevent loss of information from ReLU. Compared to the conventional CNN models, the main building block for EfficientNet is MBConv, an inverted bottleneck conv, known initially as MobileNetV2. Before the EfficientNets came along, the most common way to scale up ConvNets was either by one of three dimensions - depth (number of layers), width (number of channels), or image resolution (image size). EfficientNets, on the other hand, perform Compound Scaling - that is, scale all three dimensions while maintaining a balance between all dimensions of the network. We use this trained model for further analysis using Mask-Unmask classification frameworks. 

To calculate the fit score of the appropriate region covering the nose and mouth region of the face, we used a semantic segmentation based model (Figure 1) and defined a fit score as shown in (Equation 1). Using the data from Section 2.1, we trained a U-net\cite{ronneberger_2015_unet} based model for segmenting images of faces into the mask and unmasked regions. The model uses ResNet 32 and 50 encoders, pre-trained on ImageNet data \cite{deng_2009_imagenet}. The Layers were trained progressively using cyclical learning rates \cite{smith_2018_a}. Different model variations are tried using different encoders, and input image sizes are shown in (Table 13). Using the output of this model (True Positive(TP) + False Positive(FP)) and the nose-mouth region's facial landmarks, we calculated the fit score of an image of a face. The fit score is the percentage of the area of ROI covered by the mask on the face (Figure \ref{fig:fig5}). We employed the Fit score analysis framework on the BLM posts to get an insight into how well people wore masks in groups during large evens like protests. City wise analysis was done to observe mask fit differences across major states of protest.

$$Fit Score = \frac{|(TP \cup FP) \cap ROI|}{|ROI} * 100 \label{fit_score}$$

\subsection{Statistical Tests}
We use Mann-Kendall trend test to look for monotonic increasing trends in the daily percentage mask wearers.

We also use Pearson’s correlation, Spearman’s rank-sum correlation and Welch’s t-test to perform our analysis. Although Pearson’s correlation assumes normal distribution for both variables \cite{rovetta_2020_raiders}, it has been shown to reveal hidden correlations even when data are not normally distributed \cite{welch_1947_the}. 

We performed Pearson and Spearman correlation  to decide whether the value of the correlation coefficient  r  between lagged COVID-19 cases and daily percentage of people wearing masks is significantly different from zero at threshold of P<0.01[\cite{greenland_2016_statistical}]. 

We then conducted a Welch’s t-test to assess whether the daily posting is significantly affected by stay-at-home laws. Welch’s t-test requires normal distributions as a prerequisite, but since we are comparing mean values and our underlying series length is large(>30), this assumption can be bypassed \cite{kwak_2017_central}.
We perform the Welch’s t-test to test the following hypothesis for the six cities and calculate the P-Values with alpha =0.01 :

H0 : $\mu_0 = \mu_1$ ( The daily percentage of group posting is not affected by the stay-at-home laws)
\newline
H1 : $\mu_0 \neq \mu_1$ (The daily percentage of group posting is affected by the stay-at-home laws.)

where $\mu_0,\mu_1$ are the mean percentages of daily group posting.

In addition, we perform the Welch’s t-test to test the following hypothesis for Boston, Minneapolis and New York City (mask mandate dates for other three states did not lie in out chosen  timeframe) and calculate the P-Values with alpha =0.01: 
\newline
H0 : $\mu_0 = \mu_1$  (The daily percentage of masked faces is not affected by the mask mandates)
\newline
H1 : $\mu_0 \neq \mu_1$, (The daily percentage of masked faces is affected by the mask mandates)

where $\mu_0,\mu_1$ are the  daily  mean percentages of masked faces.
We calculate the associated P-value for significance testing, with alpha =0.01.

\section{Results}
\begin{figure}[ht!]
	\centering
	\includegraphics[width=0.95\linewidth]{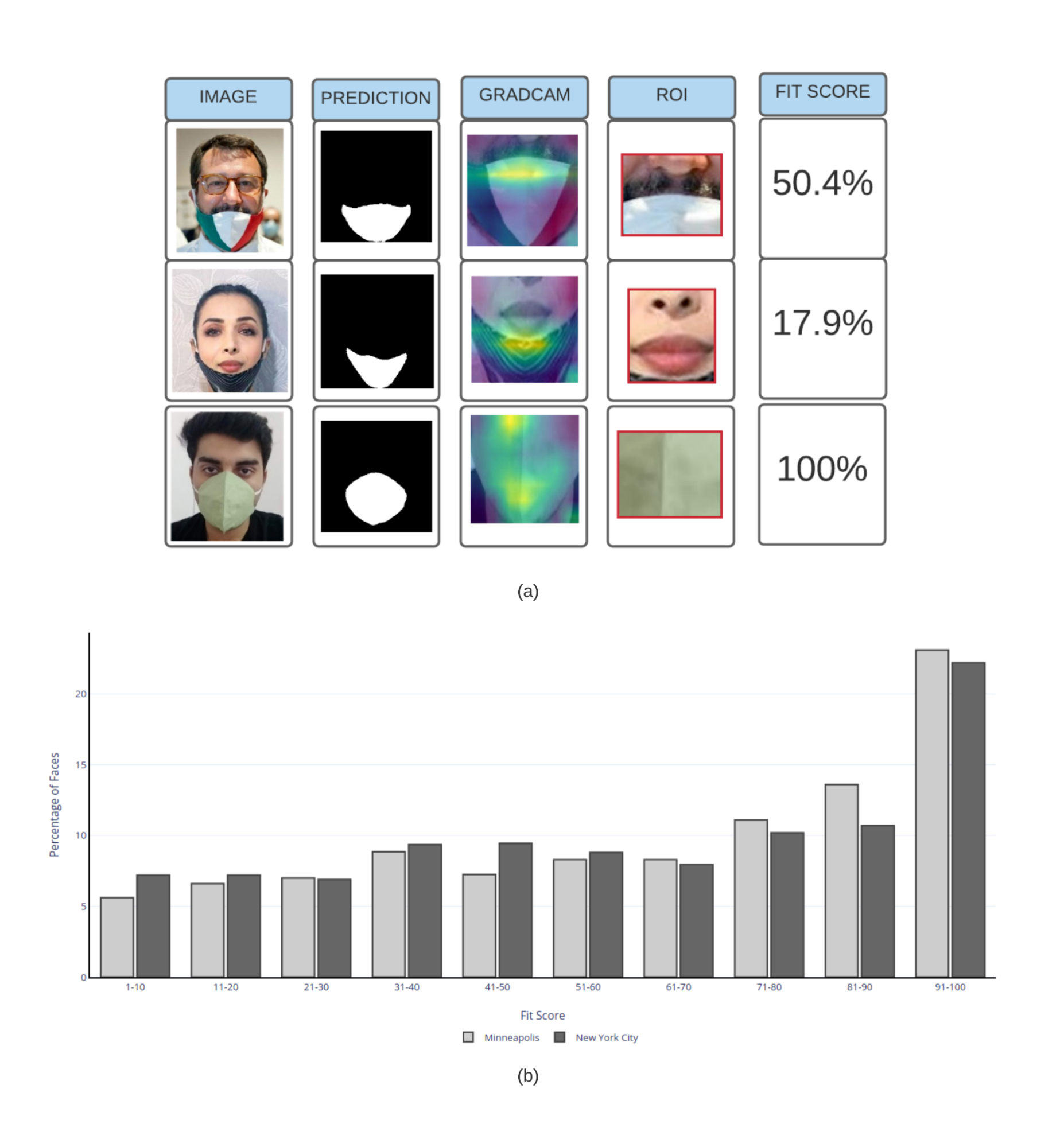}
	\caption {(a) GradCam Analysis, showing the activation of different regions on the jaw in the classification model. (b) Percentage of Faces v/s Fit Score for New York and Minneapolis for BLM posts between May 25, 2020, and July 15, 2020. Around a quarter of the masked people were wearing the mask with a fit score of greater than equal to 90 \%. A total of 11,214 posts were analyzed.}
	
	\label{fig:fig2}
\end{figure}

To visually inspect what the trained EfficientNet B0 in mask-unmask classifier had learned, we implemented GradCam \cite{selvaraju_2017_gradcam} on the network. GradCam assesses which parts of the input image have the highest activation values, given a target class. In this case, we passed the jaw region (ROI) after facial landmark detection as the input image to the GradCam network for three examples, two masked and one unmasked. . The corresponding activation maps, mask predictions and fitscores are shown in (Figure \ref{fig:fig2}(a)).  
We also inspected the segmentation model on the corpus of BLM-posts collected from social media. Figure \ref{fig:fig2}(b) shows the distribution of the fit scores, for people wearing masks in BLM-posts. Around only 45\% of the detected faces had a fit score of greater than 80\%. This means that the remaining 55\% had some major part of their nose-mouth region not covered.

\begin{table}[]
\begin{tabular}{|l|l|l|l|l|}
\hline
City          & \textit{Total Collected posts} & \textit{Faces Detected} & \textit{Masks Detected} & Percentage of Faces with Masks \\ \hline
              &                                &                         &                         &                                \\ \hline
New York City & 245,677                        & 200,089                 & 25,413                  & 12.70                          \\ \hline
Dallas        & 540,500                        & 444,194                 & 48,119                  & 10.83                          \\ \hline
Seattle       & 437,040                        & 312,012                 & 46,019                  & 14.75                          \\ \hline
Minneapolis   & 220,999                        & 152,822                 & 30,385                  & 19.88                          \\ \hline
New Orleans   & 315,082                        & 321,591                 & 39,420                  & 12.26                          \\ \hline
Boston        & 283,757                        & 238,770                 & 43,350                  & 18.15                          \\ \hline
Total         & 2,043,055                      & 1,669,478               & 232,706                 & 13.94                          \\ \hline
\end{tabular}
\caption{City-wise distribution of the number of detected faces, detected masks and number of Masks per Face through our frameworks.}
\end{table}

The following paragraph presents the results of experiments conducted to evaluate patterns of people wearing masks in the six cities across the selected time frame,  February 1, 2020 and May 31,  2020. A total of 1.66 million faces were detected from all the posts across the six cities. Out of which, a total of 232,706 faces had masks. (Table 4) shows the city-wise distribution of the detected faces and masks. 1.16 million posts (around 57\% of the total posts collected) had no faces, while 1.89 million (around 93\%) of the total posts had no masked faces. One or more faces were detected in 0.87 million (43\%) of the posts, of which 0.61 million(30\%) had a single face detected, and 0.26 million (13\%) had multiple faces detected. While in 0.14 million(7\%) of the total posts, one or more masked faces were detected, out of which 0.12 million(6\%) had a single masked face detected.

\begin{figure}[ht!]
	\centering
	\includegraphics[width=0.95\linewidth]{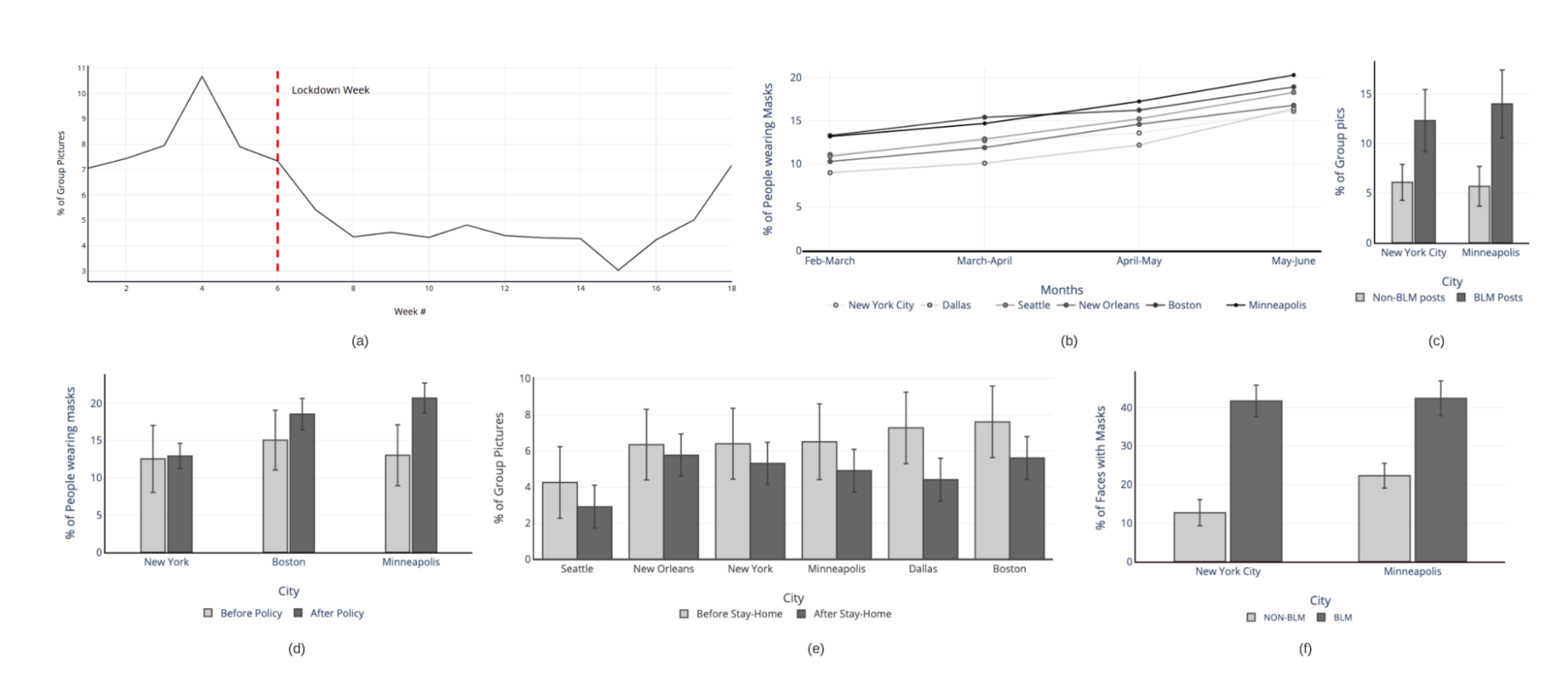}
	\caption {(a) Weekly percentages of group pictures detected from all 6 cities between February 1,  2020, and May 31, 2020. A total of 2.04 million posts were analysed. (b)  Monthly percentages of people wearing masks for each of the six cities, between February 1,  2020 and May 31, 2020. For each city, the dataset was divided into months, and the percentages of people wearing masks were computed. (c) Percentage of group pictures v/s city for BLM and non-BLM posts between May 25, 2020 and July 15, 2020. More group pictures were observed in the BLM posts than the Non-BLM posts.  A total of 192,854 posts analysed.  (d) Average daily Percentage of people wearing masks, before and after mask-wearing guidelines for New York, Boston, and Minneapolis, between February 1,  2020, and May 31, 2020. There is an increase in the percentage of people wearing masks post the mask-wearing mandates. A total of 750,433 posts were analysed. (e) Average daily Percentage of Group pictures before and after Stay-at-Home laws for the six cities, between February 1,  2020, and May 31, 2020. A total of 2.04 million posts were analysed.  (f) Percentage of people wearing masks in groups for BLM and non-BLM posts between May 25, 2020, and July 15, 2020. A total of 27,789 posts were analysed.}
	
	\label{fig:fig3}
\end{figure}

There is a decrease in the group posting after the lockdown week. The average value of percentage group pictures dropped from 8.05\% to 4.65\%. A sudden spike in the group posting is observed around week 15 (Figure \ref{fig:fig3}a). 
A general increasing trend in the percentage of people wearing masks for all six cities was observed. Mann-Kendall’s trend test  showed a significant positive trend in daily percentage of mask wearers for all 6 cities(Corresponding P values can be found in Table 14).  New York City, Dallas, Seattle, New Orleans, Boston and Minneapolis observed a month-wise increase of 5\%, 7.4\%, 7.4\%, 6.5\%, 5.6\% and 7.1\% respectively between February 2020 and May 2020 (Figure \ref{fig:fig3}b).
As shown in (Figure \ref{fig:fig3}c) the difference between BLM and NON-BLM posts in group pictures was 6.2\% and 8.3\% for New York City and Minneapolis respectively, whereas the difference between BLM and NON-BLM posts in Percentage of masked faces in group pictures was 29\% and 20.1\% for New York City and Minneapolis respectively ( Figure \ref{fig:fig3}f).

(Figure \ref{fig:fig3}d) shows the average daily percentage of people wearing masks before and after the state mask mandates were applied for the three cities that implemented these mandates within our selected time range. The time of issues of stay-at-home laws and numbers for each of the 6 cities can be found in the Appendix. Boston, Minneapolis, and New York City saw an increase of 3\%, 7.4\%, and 1\% respectively after the application of mask-mandates. The average daily percentages of people wearing masks before and after the state mask mandates were applied, are statistically different from one another for Boston and Minneapolis with alpha = 0.01, while the difference is not significant for New York City. Table 7 in appendix shows the results of Welch's t-test for the 3 cities.
(Figure \ref{fig:fig3}e) shows the average daily percentage of group pictures being posted for each of the six cities before and after the stay-at-home laws were put into place. The time of issues of stay-at-home laws and numbers  for each of the 6 cities can be found in the appendix.  Boston, Minneapolis, New Orleans, Dallas and New York City saw a decrease of 2\%,1.6\%, 0.58\%, 2.8\%, 1.3\% and 1\% respectively. The average daily percentages of group posting before and after the stay-at-home laws were applied, are statistically different from one another for all the 6 cities with alpha set at 0.01. Table 8 in appendix shows the results of Welch's t-test for the 6 cities.

\section{Discussion}

\subsection{Principal Results}
The Covid-19 pandemic has given the entire research community and governments a chance to reflect on what kind of system needs to be placed to handle such catastrophes. A renewed focus is emerging in infodemiology \cite{eysenbach_2009_infodemiology}, especially leveraging mass surveillance data \cite{chu_2020_early}\cite{sharma_2020_use}. Location tracking \cite{saran_2020_review}, periodic self-checks, and image recognition systems have been deployed by many governments \cite{amit_2020_masssurveillance,zhu_2020_coronavirus,kharpal_2020_use,kharpal_2020_coronavirus} to get a handle on the pulse of the pandemic in their states. This study suggests another such handle, which can be applied to certain demographics to achieve similar near-real-time tracking of the pandemic's spread. Instagram and other social media platforms have been very successful in tracking the number of visitations to public places \cite{tenkanen_2017_instagram}. In the context of Covid-19, public places are the focal points for the spread of viruses. It has been well-documented that face masks and social distancing are the two most effective non-pharmaceutical interventions (NPIs) to curb the spread of Covid-19 \cite{rader_2021_maskwearing}. However, the use of masks and effective social distancing are often self-reported \cite{maloney_2020_the}, without any proof to corroborate the claims. Models built on image data with location data \cite{li_2020_mask} can be a powerful tool for the authorities to keep track of the pandemic's pulse.

\begin{figure}[ht!]
	\centering
	\includegraphics[width=0.95\linewidth]{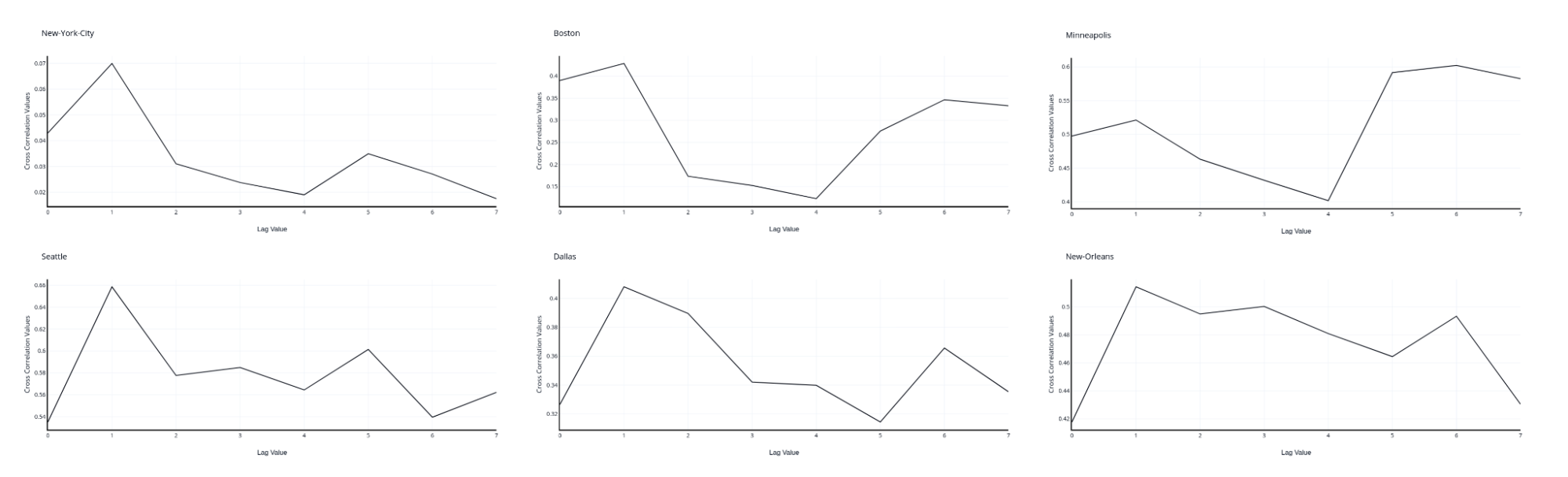}
	\caption {Pearson’s Correlation between daily lagged cumulative cases and percentage of masked photos, between February 1,  2020 and May 31, 2020. The length of the series was 120. The lag was selected between 0 and 7 based on the highest correlation value. }
	
	\label{fig:fig4}
\end{figure}

An overall decrease in the group posting is found as the pandemic grew and lockdowns were put in place. These group pictures posted online can be used as an estimator for the percentage of people spending time in groups. A sudden spike in the group posting was found around May 16, 2020 -May 22, 2020 (week 15 of our timeline). This can be linked to the easing lockdown restrictions through weeks 13,14, and 15  \cite{tribune_2020_texas,press_malls,inslee_2020_inslee,gov,holshue_2020_first}  . As the pandemic grew, the percentage of mask-wearers also saw a significant increasing trend for all the six cities, through the months, suggesting more people started wearing masks as the pandemic grew. A significant positive Spearman’s and Pearson’s correlation is found for all the cities between the daily COVID-19 cases and the percentage of masked wearers except for New York City, with alpha = 0.01. The maximum correlation was found to be with lag as one, for all cities, except Minneapolis as seen in Figure \ref{fig:fig4}.  As seen through social media the stay-at-home state policies were successful as a significant decrease in group posting was observed on the application of stay-at-home laws for all six cities that were analysed. After the mask-wearing mandates were applied, a significant increase in the percentage of mask-wearers was seen in Boston and Minneapolis, and a slight increase was observed for New York City. These results indicate adherence to non-pharmaceutical interventions for the six cities, with varying percentages of changes and effects. The trends of increasing mask-wearing with the pandemic and positive changes upon mask mandates corroborate with the self-reported number based survey methods in the USA \cite{rader_2021_maskwearing}. Although a significant increase was seen in the percentages, the growth could have increased separately from the mandates. With an insignificant increase seen for New York City,  supplemental public health interventions can be applied to maximize the adoption of such methods.

A large difference is observed in the percentage of group pictures between the posts that talked about the BLM protests. This difference can be explained by the huge collection of people in protests, with lack of social distancing measures causing a high percentage of posts being group pictures. On the contrary, the mask-wearing was found to be much higher for BLM related posts as compared to Non-BLM posts. The mask-fit score distribution of the protestors show that only 35\% of the mask-wearers had more than 80\% of their nose-mouth region covered. This indicates that, while social distancing measures were not followed properly, due to the nature of such large gatherings, protestors were more likely to wear a mask than the general public, but only a small percentage covered their faces properly as seen through social media. 
Models built on image data with location data can be a powerful tool for the authorities to keep track of the pandemic's pulse.  This study provides a new method for governments and organisations to indirectly monitor policy decisions. 

\subsection{Limitations}
The images present in datasets, VAMA-C and VAMA-S, are high definition and in RGB mode. If the trained models have to be deployed in a new setting like CCTV feed, certain image augmentation techniques like grayscaling and rescaling might be needed for fine-tuning the model. 
For the analysis, we choose the image data from six major US cities by population and correlate it with state-wide Covid-19 cases. Since these cities are some of the most populated cities of the respective states, it is reasonable to assume that they will be the hot-beds of the Covid-19 spread in that state. 
The analysis has been conducted using images fetched from the social-media platform Instagram. The authors understand that these images might not be a representation of the entire city population \cite{blank_2017_representativeness}. However, they do represent a wide demographic of internet users \cite{boy_2016_how}. With recent studies conducted on geo-tagged text data present in Instagram posts \cite{rovetta_2020_global}, our assumption of a fair population representation in the Instagram data might not be too far-fetched. Among the mask wearers the celebrity posts(posts with likes greater than 10000) contributed to only 0.2\% of the total posts which contained at least one mask, which shows that major contribution to the above collected data is through the general population. However, the authors do acknowledge that capturing meta-data for users while performing similar studies might yield a conclusive answer to the question of fair representation. Capturing and using meta-data can be a future work, building on these results. 

\subsection{Future Works}
An addition to the modeling pipeline would be an indoor/ outdoor environment detector, similar to Zhou, et al. \cite{zhou_2018_places}. Another addition to the analysis could be selectively looking at specific user activity who are deemed as "influencers" on the network. Their activity on the network can be analysed in conjunction with the activity of their followers. This can help determine the role of social networks and the power of certain influential nodes in that network over other people's behavior during critical times such as a pandemic. Dynamic location relationships present in mobility data \cite{covid19} can be used to further understand the pandemic's spread with higher location precision and even recognise malevolent actors in the system. A natural extension of this work is its replication across different social media platforms like Twitter, Facebook, Baidu.

\subsection{Conclusion}
Models built on image data with location data can be a powerful tool for the authorities to keep track of the pandemic's pulse. This study examines 2.04 million posts collected from six cities from the USA between February 1, 2020, and May 31, 2020, for adherence to mask-wearing and social distancing as seen through social media.
This study finds a general increasing trend in mask-wearing and a decreasing trend of group pictures as the pandemic grew. The stay-at-home laws showed a significant drop in group posting for all six cities, while the mask mandates showed a significant increase in mask wearing for two of the three cities analysed. Although these results suggest an upward trend in adoption of preventive methods, a large portion of non-adopters seen online, indicate a need for supplemental measures to increase the effectiveness of such methods.
Posts related to protests were found to capture the lack of attention given to safety measures through high percentage of detected group pictures, and incorrect mask-wearing. The methodology used provides a directional indication of how government policies can be indirectly monitored. These findings can help governments and other organisations as indicators for a successful implementation of non-pharmaceutical interventions for the pandemic.

\subsection{FAIR}
The gathered data consists of publicly available information about a social network. Gathering and examining of this data provided significant insights into public health recommendations. Our dataset also conforms to the Fairness, Accessibility, Interoperability, and Reusability of the collected dataset (FAIR) principles. In particular, the datasets used for training the models, are  “findable”, as it is shared publicly at \cite{precog}. This dataset is also “accessible”, given the format used (.png) is popular for data transfer and storage. This file format also makes the data “interoperable” and  “reusable”, given that most programming languages and softwares have libraries to process image files. The data was collected through public API endpoints of Instagram. The data we collected was stored in a central server with restricted access and firewall protection.

%
\bibliographystyle{spmpsci}
\bibliography{main}

\newpage
\newpage
\section*{Appendix}

\begin{table}[!h]
\begin{tabular}{|l|l|l|l|}
\hline
City          & \textit{Max Lag} & \textit{Correlation Value} & \textit{P Values} \\ \hline
New York City & 1                & 0.07                       & .50               \\ \hline
Boston        & 1                & 0.42                       & \textless{}.001   \\ \hline
Minneapolis   & 6                & 0.60                       & \textless{}.001   \\ \hline
Seattle       & 1                & 0.65                       & \textless{}.001   \\ \hline
Dallas        & 1                & 0.40                       & \textless{}.001   \\ \hline
\end{tabular}
\caption{Pearson’s correlation coefficient and P-values, between Lagged cumulative COVID-19 cases and daily percentages of people wearing masks. }
\end{table}

\begin{table}[!h]
\begin{tabular}{|l|l|l|l|}
\hline
City          & \textit{Max Lag} & \textit{Correlation Value} & \textit{P Values} \\ \hline
New York City & 1                & 0.05                       & .62               \\ \hline
Boston        & 1                & 0.60                       & \textless{}.001   \\ \hline
Minneapolis   & 2                & 0.42                       & \textless{}.001   \\ \hline
Seattle       & 1                & 0.71                       & \textless{}.001   \\ \hline
Dallas        & 1                & 0.47                       & \textless{}.001   \\ \hline
New Orleans   & 1                & 0.54                       & \textless{}.001   \\ \hline
\end{tabular}
\caption{Spearman’s correlation coefficient and P values, between Lagged cumulative COVID-19 cases and daily percentages of people wearing masks.}
\end{table}

\begin{table}[]
\begin{tabular}{|l|l|l|}
\hline
City          & \textit{t statistic} & \textit{P Values} \\ \hline
New York City & -4.35                & \textless{}.001   \\ \hline
Boston        & -6.0                 & \textless{}.001   \\ \hline
Minneapolis   & -2.75                & .008              \\ \hline
Seattle       & -6.53                & \textless{}.001   \\ \hline
Dallas        & -8.86                & \textless{}.001   \\ \hline
New Orleans   & -8.74                & \textless{}.001   \\ \hline
\end{tabular}
\caption{Welch’s t-test statistic, P values to test for equal means before and after application of stay-at-home laws.}
\end{table}

\begin{table}[]
\begin{tabular}{|l|l|l|}
\hline
City        & \textit{Date}  & \textit{Comment}                                                                                     \\ \hline
Boston      & March 23,2020  & Boston Department of Public Health issued a two-week stay-at-home advisory on \\ & & March 23, 2020 \cite{a2020_coronavirus} \\ \hline
Minneapolis & March 27, 2020 & For Minneapolis, stay-at-home advisory came into effect on March 27, 2020 \cite{a2020_governor}                   \\ \hline
New Orleans & March 20, 2020 & New Orleans enacted stay-at-home orders starting March 20, 2020 \cite{a2020_mayor}                              \\ \hline
Dallas      & March 23, 2020 & Dallas enacted stay-at-home orders starting  March 23, 2020 \cite{pugh_2020_dallas}.                                \\ \hline
Seattle     & March 23, 2020 & Seattle enacted stay-at-home orders starting  May 31, 2020 \cite{a2020_washington}.                                  \\ \hline
New York    & March 20, 2020 & In New York State state-wide stay-at-home order was declared on March 20, 2020 \cite{a2020_governer1}               \\ \hline
\end{tabular}
\caption{Dates at which stay at home guidelines were enacted by the respective state governments.}
\end{table}

\begin{table}[]
\begin{tabular}{|l|l|p{6cm}|}
\hline
City           & \textit{Date}  & \textit{Comment}                                                                                                               \\ \hline
Boston         & May 6, 2020    & In Boston, residents were asked to wear a mask in public places starting May 6, 2020 {[}2{]}                                   \\ \hline
Minneapolis    & April 30, 2020 & Mask wearing guidelines were put in place on April 30, 2020 for Minneapolis, encouraging people to wear face coverings \cite{a2020_emergency} \\ \hline
New York State & April 15, 2020 & In New York State, residents were asked to wear masks in public starting April 15, 2020 \cite{a2020_amid}                                \\ \hline
\end{tabular}
\caption{Dates at which mask mandates were enacted by the respective state governments.}
\end{table}

\begin{table}[]
\begin{tabular}{|l|l|l|}
\hline
City          & \textit{t statistic} & \textit{P Values} \\ \hline
New York City & 1.68                 & .09               \\ \hline
Boston        & 4.19                 & \textless{}.001   \\ \hline
Minneapolis   & 3.64                 & \textless{}.001   \\ \hline
\end{tabular}
\caption{Welch’s t-test statistic, P values to test for equal means before and after application of mask mandates.}
\end{table}

\begin{table}[]
\begin{tabular}{|l|l|l|l|l|l|l|}
\hline
Time Period & New York City & \textit{Dallas} & Seattle      & New Orleans  & Boston      & Minneapolis \\ \hline
Feb-March   & 4847/45301    & 10369/112882    & 6211/56650   & 13537/131390 & 9863/73506  & 5373/36741  \\ \hline
March-April & 5983/49869    & 11438/113014    & 9105/70312   & 9305/80011   & 13398/86987 & 6734/42279  \\ \hline
April-May   & 7545/58310    & 24878/209186    & 9958/65373   & 7083/49809   & 600/3649    & 2869/16139  \\ \hline
May-June    & 7038/46609    & 1438/9112       & 20745/119677 & 9545/60381   & 16280/75678 & 12183/49777 \\ \hline
\end{tabular}
\caption{Underlying n/N values for (Figure \ref{fig:fig3}(b))}
\end{table}

\begin{table}[]
\begin{tabular}{|p{2cm}|p{4cm}|p{2cm}|p{4cm}|p{2cm}|}
\hline
City         & Sum of daily percentages/total days before mandate & \textit{Std Dev} & Sum of daily percentages/total days after mandate & Std Dev \\ \hline
Seattle      & 216.5/51                                           & 1.98             & 200.1/69                                          & 1.19    \\ \hline
New Orleans  & 304.32/48                                          & 1.96             & 414.72/72                                         & 1.17    \\ \hline
New York     & 306.72/48                                          & 1.96             & 381.6/72                                          & 1.17    \\ \hline
Minneapolis- & 351.45/55                                          & 2.1              & 344.5/65                                          & 1.19    \\ \hline
Dallas       & 370.77/51                                          & 1.98             & 303.6/69                                          & 1.19    \\ \hline
Boston-      & 364.8/48                                           & 1.98             & 403.2/72                                          & 1.19    \\ \hline
\end{tabular}
\caption{Underlying n/N values for (Figure \ref{fig:fig3}(e))}
\end{table}

\begin{table}[]
\begin{tabular}{|p{2cm}|p{4cm}|p{2cm}|p{4cm}|p{2cm}|}
\hline
City           & Sum of daily percentages/total days before mandate & \textit{std Dev} & Sum of daily percentages/total days after mandate & std Dev \\ \hline
New York & 927.96/74                                          & 4.5              & 594.32/46                                         & 1.7     \\ \hline
Boston         & 1428.8/95                                          & 4.01             & 463.75/25                                         & 2.1     \\ \hline
Minneapolis    & 1159.67/89                                         & 4.1              & 952.2/46                                          & 2.01    \\ \hline
\end{tabular}
\caption{Underlying n/N values for (Figure \ref{fig:fig3})}
\end{table}

\begin{table}[]
\begin{tabular}{|l|l|l|}
\hline
Fit Score & Minneapolis & \textit{New York City} \\ \hline
1-10      & 1281/23805  & 1659/24038             \\ \hline
11-20     & 1576/23805  & 1732/24038             \\ \hline
21-30     & 1687/23805  & 1681/24038             \\ \hline
31-40     & 2104/23805  & 2254/24038             \\ \hline
41-50     & 1747/23805  & 2316/24038             \\ \hline
51-60     & 1981/23805  & 2130/24038             \\ \hline
61-70     & 1995/23805  & 1939/24038             \\ \hline
71-80     & 2632/23805  & 2466/24038             \\ \hline
81-90     & 3233/23805  & 2559/24038             \\ \hline
91-100    & 5569/23805  & 5302/24038             \\ \hline
\end{tabular}
\caption{Underlying n/N values for (Figure \ref{fig:fig2}(b))}
\end{table}

\begin{table}[]
\begin{tabular}{|l|l|l|l|}
\hline
Model           & No. of Parameters & \textit{Recall} & Accuracy    \\ \hline
MobileNet V2    & 3,538,984         & 0.90 ± 0.01     & 0.94 ± 0.01 \\ \hline
Nas Net Mobile  & 5,326,716         & 0.86 ± 0.01     & 0.91 ± 0.01 \\ \hline
EffecientNet B0 & 5,330,571         & 0.97 ± 0.01     & 0.98 ± 0.00 \\ \hline
EffecientNet B1 & 7,856,239         & 0.97 ± 0.01     & 0.98 ± 0.00 \\ \hline
DenseNet121     & 8,062,504         & 0.90 ± 0.01     & 0.94 ± 0.01 \\ \hline
EffecientNet B2 & 9,177,569         & 0.97 ± 0.01     & 0.98 ± 0.00 \\ \hline
\end{tabular}
\caption{Face Mask Detection Model Results. EffecientNet B0 is used as the model for face mask detection due to its superior performance and smaller model size. The precision for each model was found to be 1.00 $\pm$ 0.01}
\end{table}

\begin{table}[]
\begin{tabular}{|l|l|l|l|l|}
\hline
Image Size & Encoder   & \textit{Recall} & Accuracy & IOU  \\ \hline
100 x 100  & Resnet 50 & 0.96            & 0.98     & 0.95 \\ \hline
150 x 150  & Resnet 50 & 0.96            & 0.98     & 0.95 \\ \hline
224 x 224  & Resnet 32 & 0.96            & 0.98     & 0.95 \\ \hline
\end{tabular}
\caption{Face Mask Fit Analyser Model Results. IOU refers to the Intersection Over Union (IOU) score.}
\end{table}

\begin{table}[]
\begin{tabular}{|l|l|l|}
\hline
City          & Normalized test statistic (z) & \textit{P Values} \\ \hline
New York City & 4.1                           & \textless{}.001   \\ \hline
Boston        & 6.34                          & \textless{}.001   \\ \hline
Minneapolis   & 5.65                          & \textless{}.001   \\ \hline
Seattle       & 8.72                          & \textless{}.001   \\ \hline
Dallas        & 8.22                          & \textless{}.001   \\ \hline
New Orleans   & 8.47                          & \textless{}.001   \\ \hline
\end{tabular}
\caption{Test statistic and P Values for  Mann-Kendall trend test for daily percentage masked wearers in six cities}
\end{table}

\begin{figure}[ht!]
	\centering
	\includegraphics[width=0.55\linewidth]{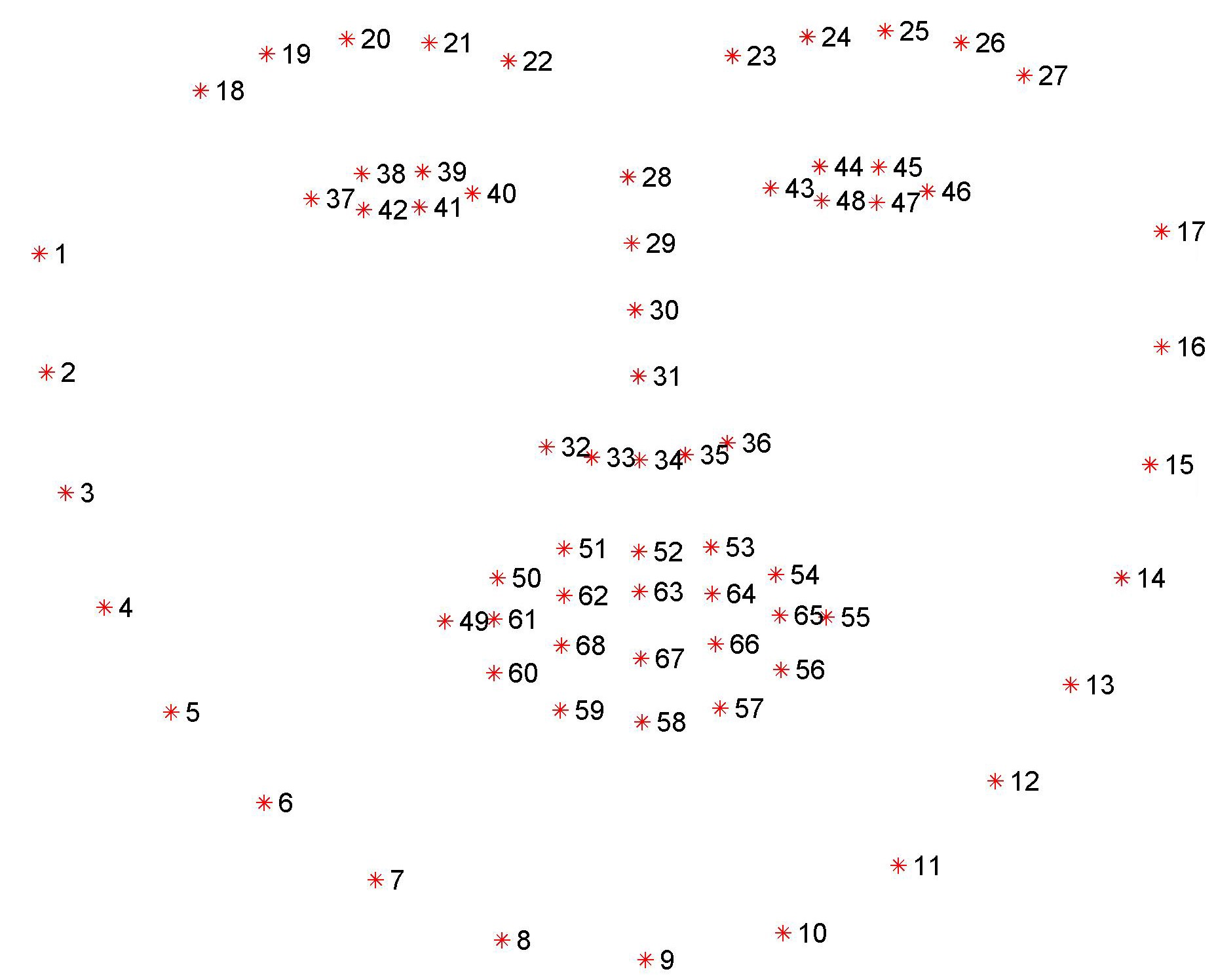}
	\caption {Facial Landmarks detected on a face using DLib. }
	
	\label{fig:fig5}
\end{figure}


\end{document}